\documentclass[11pt]{article}
\usepackage{mathpazo}

\usepackage{microtype}
\usepackage{subcaption}
\usepackage{graphicx}
\usepackage{booktabs} %
\usepackage{xspace}
\usepackage{xcolor}         %
\usepackage[margin=1in]{geometry}
\usepackage{enumitem}
\usepackage{wrapfig}

\usepackage[backend=biber,style=alphabetic,natbib=true,maxcitenames=2,minalphanames=3]{biblatex}
\newcommand{\ffcv}{\texttt{FFCV}\xspace}

\usepackage[colorlinks=true,linkcolor=blue,citecolor=blue]{hyperref}

\usepackage{comment}

\title{FFCV: Accelerating Training by Removing Data Bottlenecks}
\author{
    Guillaume Leclerc\footnote{Equal contribution.} \\
    \texttt{leclerc@mit.edu} \\
    MIT
    \and
    Andrew Ilyas\footnotemark[1] \\
    \texttt{ailyas@mit.edu} \\
    MIT
    \and 
    Logan Engstrom\footnotemark[1] \\
    \texttt{engstrom@mit.edu} \\
    MIT
    \and
    Sung Min Park \\
    \texttt{sp765@mit.edu} \\
    MIT
    \and
    Hadi Salman \\
    \texttt{hady@mit.edu} \\
    MIT
    \and
    Aleksander Madry \\
    \texttt{madry@mit.edu} \\
    MIT
}
\date{}
\begin{document}

\maketitle

\begin{abstract}
  We present \ffcv, a library for easy and fast machine learning model training.
  \ffcv speeds up model training by eliminating (often subtle) data bottlenecks
  from the training process. 
  In
  particular, we combine techniques such as an efficient file storage format,
  caching, data pre-loading, asynchronous data transfer, and just-in-time
  compilation to (a) make data loading and transfer significantly more
  efficient, ensuring that GPUs can reach full utilization; and (b) offload as
  much data processing as possible to the CPU asynchronously, freeing GPU
  cycles for training. Using \ffcv, we train ResNet-18 and
  ResNet-50 on the ImageNet dataset with competitive tradeoff between
  accuracy and training time.
  For example, we are able to train an ImageNet ResNet-50 model to 75\% in only 20 mins on a single machine.
  We demonstrate \ffcv's performance, ease-of-use,
  extensibility, and ability to adapt to resource constraints through several
  case studies. Detailed installation instructions, documentation, and Slack support channel are available at \url{https://ffcv.io/}. 
\end{abstract}

\section{Introduction}
What is the limiting factor in faster model training? Hint: it is not always the
GPUs.
When training a machine learning model,
the life cycle of an individual example spans three stages: reading the example into memory,
processing the example in memory, and finally updating model parameters with the
example on GPU (e.g. by calculating and then following the gradient). 
The
stage with the lowest throughput determines the overall learning system's
throughput.

\begin{figure}[!b]
    \centering
    \includegraphics[width=\linewidth]{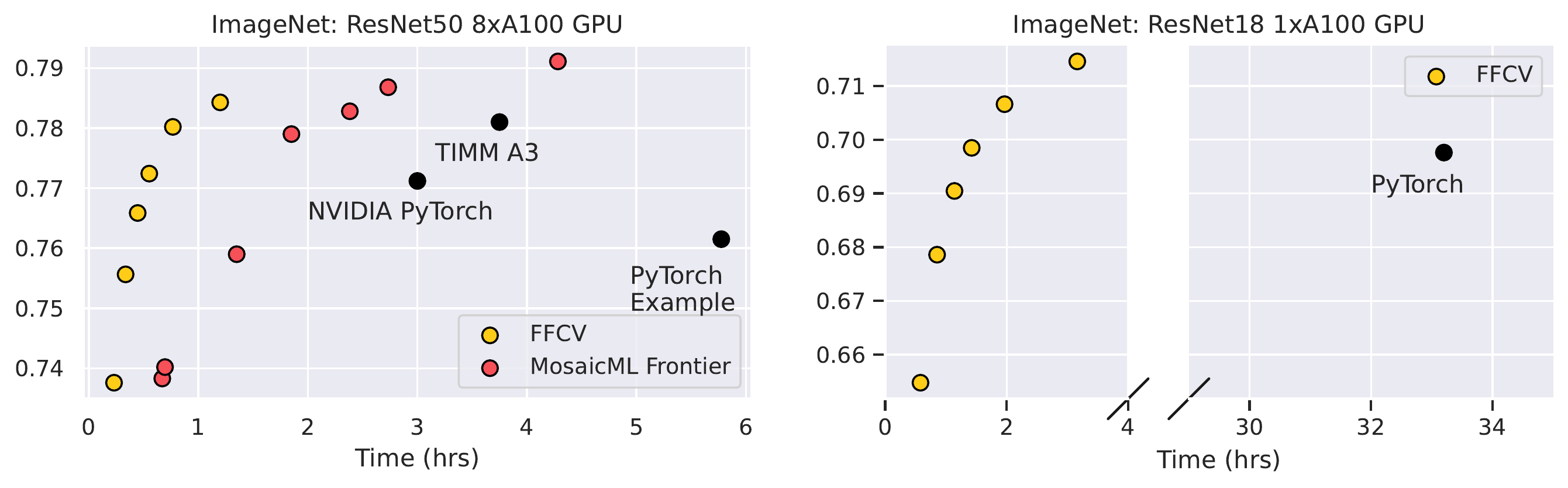}
    \caption{
    Accuracy vs. training time when training a ResNet-50 on 8 A100s. \ffcv{}
    achieves competitive accuracy/training time trade-offs compared to standard
    baselines. As an example, we can train ImageNet to 75\% accuracy in less
    than 20 minutes on a single machine. 
    In the plot on the left, the red dots correspond to the Pareto frontier of
    models achievable with the MosaicML fast training solution as of February
    2022---since then, MosaicML has integrated FFCV into their training
    pipelines.
    }
    \label{fig:scatterplot}
\end{figure}

Our investigations (and others'~\citep{mohan2021analyzing}) show
that in practice the limiting factor is often not computing updates, but
rather the data reading and data processing stages. Indeed, in standard training
setups, \textit{the GPUs can spend a majority of cycles just waiting for inputs
to process!}

To better saturate GPUs and thereby increase training
throughput, we present \ffcv{}, a system designed to reduce data
loading and processing bottlenecks while remaining simple to use. 
\ffcv{} operates in two successive
stages: \textit{preprocessing} and \textit{train-time loading}. In the first stage, \ffcv{} preprocesses the dataset into a format
more amenable to high-throughput loading. Then, in the
\textit{train-time loading} stage, \ffcv{}'s data loader replaces the original
learning system's data loader without requiring \textit{any}
other implementation changes.

Together, \ffcv{} data preprocessing and the \ffcv{} data loader can drastically
increase training speeds without any learning algorithm modifications.
To demonstrate, we train machine learning models for a number of tasks much faster
than previous general purpose data loaders (\textit{e.g.,} PyTorch's default data loaders) can support.
While \ffcv{} improves performance on most GPUs, its effect is most pronounced on faster GPUs, which
require higher throughput data loading to saturate available compute capacity.
We expect \ffcv{} will only increase in utility as new GPUs become faster.

\paragraph{Contributions.} 
\vskip -0.2cm We introduce \ffcv{}, a drop-in, general purpose training system for
high throughput data loading. Using FFCV requires \textit{no} algorithmic
changes, and involves a nearly identical API to standard data loading systems
(e.g., the default PyTorch~\citep{paszke2019pytorch} data loader). \ffcv{}
automatically handles the necessary data transfer, memory management, and data
conversion work that users usually manually optimize (or leave to suboptimal
defaults). \ffcv{} also replaces the default data preprocessing and augmentation
pipeline with one that is more efficient due to (a) just-in-time compilation to
machine code and (b) highly optimized memory management. We explore potential use
cases of \ffcv{} and find dramatic speed-ups:
\begin{itemize}[leftmargin=.5cm]
    \item \textbf{ImageNet Training.} We greatly improve ImageNet
    single node training throughput, achieving competitive speed-accuracy
    trade-offs (Figure \ref{fig:scatterplot}).
    \item \textbf{Bootstrapping and grid search.} We enable
    faster large-scale grid search by supporting same-machine, 
    different-GPU training without any throughput penalty.
    \item \textbf{Network filesystem-based training.} Especially in
    cloud computing environments where network file systems are commonplace,
    data reading can greatly bottleneck learning systems. \ffcv{} enables
    faster data loading in a realistic read-constrained
    environment.
    \item \textbf{Tasks beyond computer vision.} We demonstrate \ffcv{}'s
    ability to speed up almost any data loading task by using it as a drop-in
    replacement to the default PyTorch data loader in a GPU-enabled sparse
    regression solver.
\end{itemize}

\section{Identifying Bottlenecks in Training}
\label{sec:bottleneck_story}
What makes a machine learning training system ``slow'' or ``fast?''
The answer varies by task, algorithm, implementation, and
computing equipment available at train time.
Model training is best thought of as a {\em pipeline} of discrete
steps: data reading, data processing, and GPU computing that executes the
learning algorithm. 

\begin{figure}[!t]
    \centering
    \includegraphics[width=.8\linewidth]{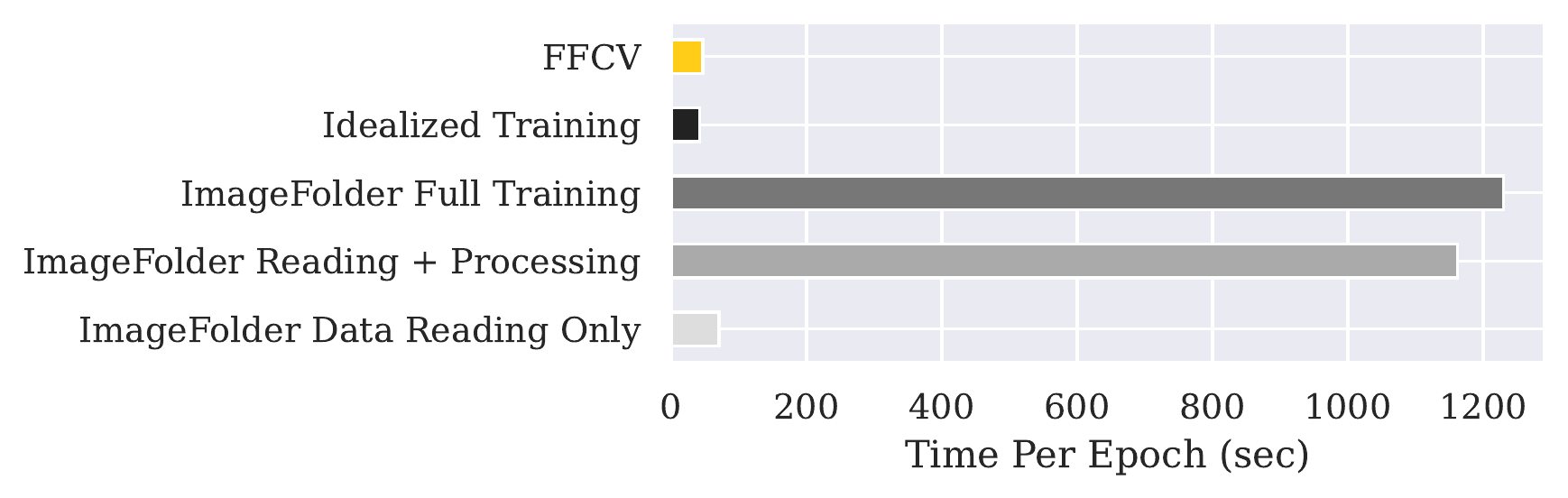}
    \caption{Time taken across stages in ImageNet training (median over three
    runs). ImageFolder refers to the default PyTorch data loader used to load
    ImageNet. We find that data loading (in particular, data processing) is the
    major bottleneck of standard training. The idealized training time, or the
    training time we would obtain with perfect data loading, is almost 30 times
    smaller than the time required to just process all the training images. In
    the top column, \ffcv{} achieves nearly ideal training time by removing the
    data reading and processing bottlenecks.}
    \label{fig:profiling_main}
\end{figure}

To understand which of these steps bottlenecks training in practice, we
study a standard task commonly used to benchmark training
speeds~\citep{coleman2017dawnbench,mattson2020mlperf}, namely
ImageNet~\citep{deng2009imagenet} training. As a specific setup, we investigate the
PyTorch ImageNet training example with the standard PyTorch ImageNet data loader,
running on a standard AWS instance for GPU-based
learning: p4d.24xlarge machines, which have 8 A100 GPUs, 96 vCPUs, and enough
RAM to fit the ImageNet training set into memory. 
We benchmark each part of the system's throughput;
Figure~\ref{fig:profiling_main} shows our results. Overall, we find 
that data loading bottlenecks this standard training setup, and, furthermore,
by fixing data loading we could achieve 30 times faster model training. Below we
explore this data loading bottleneck in further detail.

\paragraph{Data reading throughput.} \vskip -0.2cm We begin by only benchmarking
data read throughput, measuring how long the data loader takes to read the
entire dataset without performing any processing. As the machine we test on can
cache the entire ImageNet dataset into memory, the data reading step is not a
bottleneck (cf. Figure~\ref{fig:profiling_main}): it takes only 75 seconds.

\paragraph{Data processing throughput.} \vskip -0.2cm To check whether data
processing is a bottleneck, we measure how long the data loader takes to read
the entire dataset while \textit{also} performing processing: JPEG decoding,
random cropping/resizing, random flipping, and normalization. We find that
processing \textit{is} a major bottleneck: adding processing to reading greatly
increases loading time to 1200 seconds from the 70 seconds that loading alone
took (see Figure~\ref{fig:profiling_main}).

\paragraph{Full training throughput.} \vskip -0.2cm Finally, we measure the
entire system's throughput, including the learning stage. We find that adding
the learning stage does not greatly change the time taken to iterate through the
whole dataset (cf. Figure~\ref{fig:profiling_main}). The fact that our
throughput does not decrease despite adding learning on the GPUs indicates that
the data loading/processing subsystem cannot supply data fast enough to saturate
the GPUs. Indeed, as further corroboration, we simulate how fast model
throughput could be by benchmarking our training process on a fixed data vector
(here, we require no data loading). Our throughput in this idealized setting,
shown in the same figure above, is much higher, and shows that we could train
up to 30x faster with optimal data loading.

\section{Eliminating Data Bottlenecks}
\label{sec:implementation}
Now, our focus turns to the question: how can we design
a better data loading system? To maximize performance, \ffcv 
manages the entire
data management pipeline, from the file format used to store the
training data all the way to data augmentations used at training time. 
Focusing on one step of the data loading pipeline at a time, we show how
\ffcv's implementation circumvents issues in existing solutions to efficiently
load data.

\subsection{Challenge \#1: Storing a Machine Learning Dataset}
To eliminate data bottlenecks in the machine learning pipeline
we start with the data format.
There are already multiple existing file formats designed to store machine
learning datasets: the most common of these formats (and indeed, the default one
in PyTorch) is the {\em file-based format}, where one stores each
example as an individual file. 
In the context of image recognition, for example, one
saves each example as its own (typically JPEG-compressed) image file, and
uses the enclosing folder to encode the label.
The file-based approach has some advantages---most notably, users can intuitively
interact with the examples on an individual level (e.g., they can open any
training image in a standard image viewer). However, this format is not at all
optimized for performance, and comes with several fundamental drawbacks. 

\ffcv introduces its own new file format: the
\texttt{.beton} file. In the following, we discuss the different considerations
involved in designing this file format, and show that \ffcv's new file format
circumvents issues both with file-based formats as well as existing specialized
solutions (namely, WebDataset, TFRecord, and MXNet RecordIO).

\paragraph{Reducing filesystem strain.} 
\vskip -.2cm
To reduce filesystem strain, existing specialized file formats either group data
examples in shards (WebDataset) or concatenate all the data into a single file
(TFRecord, RecordIO). \ffcv adopts the latter option, but goes even further---by
organizing the dataset into pages (at the cost of some wasted space), it
eliminates random read penalties by making it easy to read data in large chunks.
\ffcv (along with TFRecord
and RecordIO) datasets are also easier to share than sharded data formats (e.g.,
WebDataset), since one only needs to transport a single file.  

\paragraph{Flexibility.} 
A data format should to be flexible enough to accommodate a wide variety of data
formats and modalities. Many existing specialized solutions 
are hyper-specialized and support only specific modalities (e.g., RecordIO
datasets can only store images with associated floating-point labels), 
while others are slightly more flexible (e.g., TFRecord and
WebDataset).
In \ffcv, we opt for {\em maximal} flexibility, and use an abstract ``Field''
class that enables users to store arbitrary data modalities (with built in support 
for vision, text, tabular, and more), and even easily extend \ffcv's capabilities
by writing data-specific 
customized encoders and decoders.

\paragraph{Searchability/Indexability.} A good data format should also natively
support {\em fast} access to only a particular subset of the dataset, whether
for the purpose of inspecting a given example, or training on a
particular subset of the training set.
Specialized data formats that only support sequential reads, however (e.g., 
TFRecord, WebDataset) are inherently unable to support such a feature.
\ffcv datasets contain a data table that hold metadata (including
indices) as well as pointers to any given sample, allowing one to 
easily filter and retrieve samples based on any predicate.

\paragraph{File structure.}
\ffcv datasets are optimized for machine learning training and offer
great performance regardless of the underlying storage method (RAM, HDDs, SSDs
or network). Each file consists of four sections. The {\em Header} contains 
general information about the dataset like the number of samples and the fields.
The {\em Data Table} 
is a small \texttt{DataFrame}-like data structure containing metadata (small fixed-width
information) about a given sample, such as e.g. the
image resolution in the image domain, or audio sample duration in the audio domain.
The {\em Heap Storage} section contains
pages (of default size $8$MB) that store either variable size information
or data that is too large for the {\em Data Table}, such as 
binary representations of images/audio examples.
Finally, the {\em Allocation Table} at the end of the file has bookkeeping
data about allocated regions in {\em Heap Storage}.

\begin{figure}
        \centering
        \includegraphics[width=.8\linewidth]{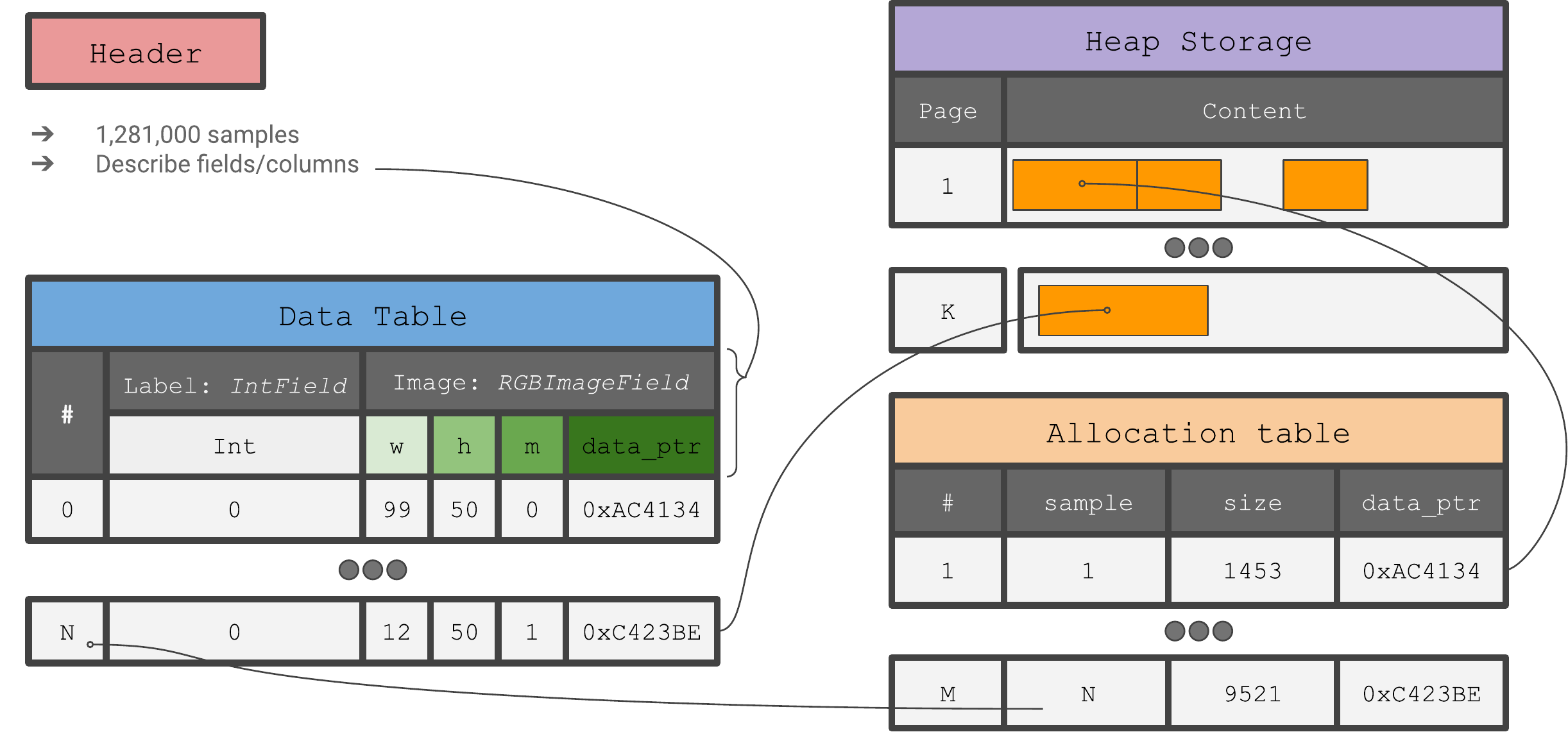}
        \caption{\label{fig:beton_imagenet_example} Structure of a
        \texttt{.beton} file used to store a simple image classification
        dataset.}
        \vskip -.3cm
\end{figure}

We show what a \texttt{.beton} file would look like for a basic image
classification task in Figure~\ref{fig:beton_imagenet_example}. 

\subsection{Challenge \#2: Efficient Data Reading}
\label{sec:challenge_2}
We now describe how \ffcv achieves
high read performance across a variety of compute environments with the \ffcv file format, ranging 
from those featuring local SSDs (with high IOPS and low latency) to large spinning disks (which suffer under random, nonsequential accesses, such as when reading many discrete
image files) to networked filesystems. \ffcv offers built-in read strategies 
optimized for high throughput across all these different scenarios.

\paragraph{Operating system caching.} 
\vskip -.2cm
For systems that can fit the dataset in random-access memory (RAM), 
\ffcv can take advantage of OS-level caching. This ensures that 
every data read after the first one
will be from RAM rather than disk, resulting in high throughput.
Beyond just simplicity, OS-level caching also allows for
multiple models training in parallel on the same dataset (i.e., when hyperparameter searching) to share the {\em same} cache without any additional memory overhead.

\paragraph{Process cache.} \vskip -.2cm
On the other hand, if the dataset is larger than the main
memory, an effective caching scheme will have to optimally cache, discard, and reload data at each epoch.
In OS-level caching---which other data loading schemes use by default---the
random access patterns induced by SGD in machine learning cause suboptimal caching
behavior. \ffcv circumvents this issue through optimized, process-level caching. 
By leveraging our knowledge of the sample order
in data loading (since we can generate this order at the beginning of
the epoch), \ffcv can preload data much earlier than the OS can.

\paragraph{Quasi-random sampling.} \vskip -.2cm
For cases where disk reads are particularly
expensive (e.g., when reading from a network drive and having insufficient RAM to
cache the dataset), \ffcv offers a {\em quasi-random loading
strategy} that can combine with the process cache strategy above to
minimize the stress on the underlying storage. 
Rather than reading examples in a
uniformly random order, the quasi-random strategy (a) allocates a buffer large
enough to fit \texttt{batch\_size} pages of the dataset; (b) samples a
permutation of all the pages of the dataset; then (c) generates a batch only
from samples in the buffer. 

WebDataset's shuffling procedure is similar to \ffcv's
quasi-random loading strategy with two crucial differences:
(1) pages in \ffcv are much smaller than WebDataset shards, leading to
significantly better randomness;~\footnote{While one can manually make WebDataset
shards smaller, this incurs a significant filesystem load.} (2) quasi-random loading in \ffcv has a constant memory
footprint, while WebDataset's footprint scales linearly in the number of
workers. For further comparison, see Appendix~\ref{app:quasirand_vs_shards}.

\subsection{Challenge \#3: Fast Data Processing}
\begin{figure}
    \centering
    \includegraphics[width=.75\linewidth]{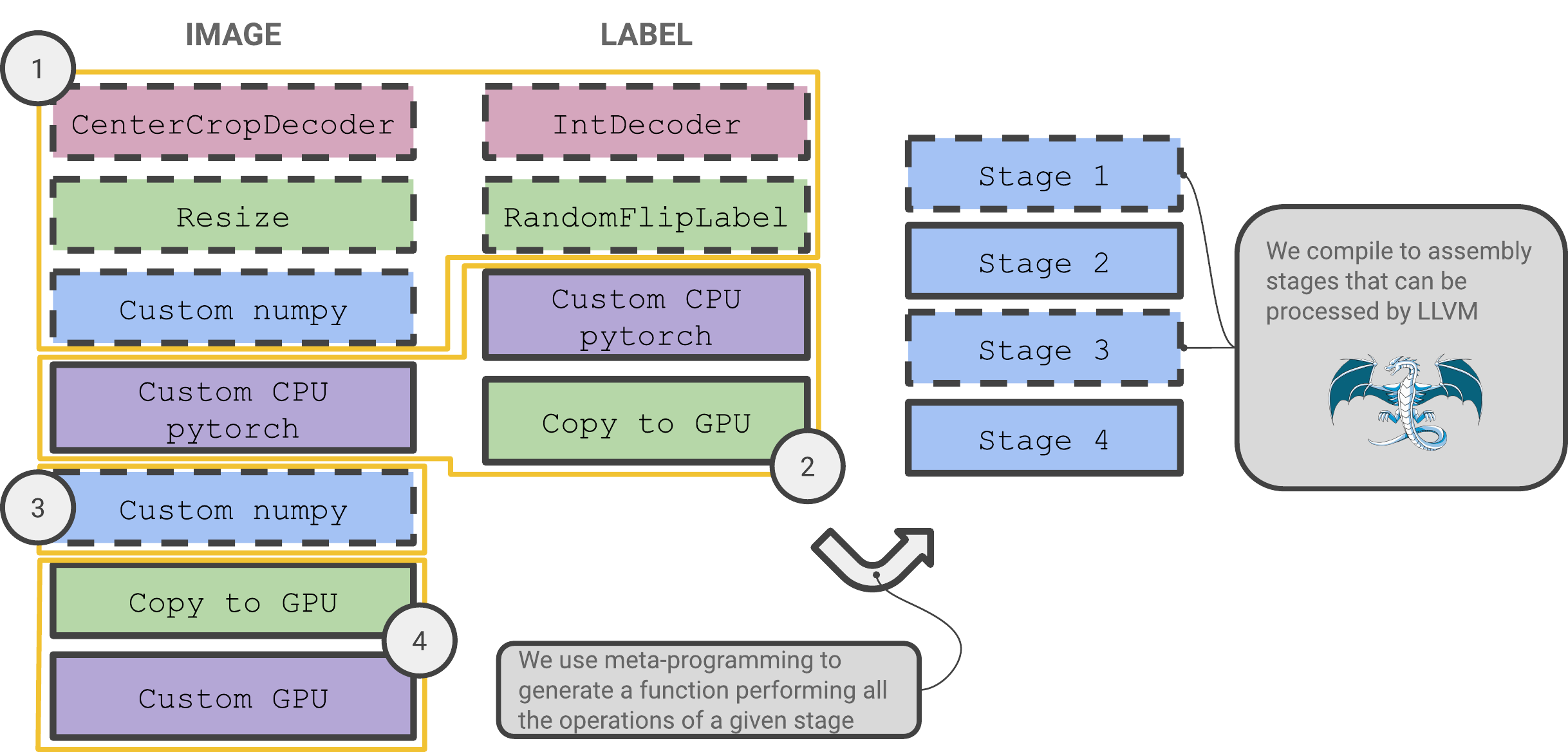}
    \caption{\label{fig:jit_example} Illustration of the procedure followed
    by \ffcv to generate the code of a complex image processing pipeline.
    Transforms are categorized together based on whether they can be JIT-ed
    (dashed, \ffcv native or numpy based user-defined augmentations) or not
    (solid, Pytorch ones and others). Groups (stages) are formed based on these categories
(1-4) by gluing each operation using meta-programming. Finally, the stages
are compiled to machine code using Numba/LLVM. }
\end{figure}

So far, we have outlined how \ffcv improves both
data storage and reading---we now turn to the {\em data processing} stage of the
machine learning pipeline.
In ML research, the data augmentation/processing pipeline requires both
efficiency (to avoid bottlenecking the entire training
process) and flexibility (to accommodate, e.g., researchers
devising their own augmentations or pre-processing techniques). 

\ffcv tries to strike a balance between these two 
objectives through a {\em just-in-time (JIT) compiled} 
data processing pipeline.
Specifically, for a small cost paid at the start of training, \ffcv 
analyzes the user-provided (Python) data processing pipeline, and automatically
compiles it to optimized {\em machine code} via the following steps:

\paragraph{Categorization.} \vskip -.2cm
Our main tool for compiling Python to machine code
is the Numba library \citep{lam2015numba}, which is by default able to compile
a large (but not complete) subset of the Python language into machine code.\footnote{We
motivate this choice and compare with other compilation systems in Appendix
\ref{app:compiler_comparison}.} We thus first {\em categorize} each element
of the data pipeline based on whether it can be automatically compiled by
Numba. (Note that all the pipeline elements that ship with \ffcv\ {\em are}
Numba-compilable, so this step is primarily to enable users to write their own
\ffcv compatible non-compilable transformations, as in cases where it is too difficult 
to write a compilable verfsion of a transformation).

\paragraph{Grouping.} 
\vskip -.2cm
After categorizing each transformation in the data pipeline, we group
together all consecutive transforms of each category into groups called
``stages'' (see Figure \ref{fig:jit_example}). This will allow us to compile
several separate pipeline elements into a single executable block of machine code.

\paragraph{Code generation.} \vskip -.2cm
Finally, using meta-programming, we generate the code necessary to
{\em fuse} each stage into a single function. Some of the stages will be then passed to 
Numba to be converted to machine code, the others remain unmodified and
run natively in Python (albeit at much lower speed than their compiled counterparts).

\paragraph{Memory pre-allocation.} \vskip -.2cm
A core tenet of \ffcv is to avoid unnecessary
memory allocation. Thus, every operation in the pipeline declares memory
requirements in advance, and memory allocation is performed {\em once} at the start
of an epoch. To let workers prepare the data while training happens, and to
absorb potential slow downs in data preparation, \ffcv relies on a circular buffer
(illustrated in Figure \ref{fig:cyclical_buffer} in Appendix~\ref{app:ommited}).

\subsection{Challenge \#4: Circumventing Data Transfer Costs}
\label{sec:challenge_4}
Since compiled machine code is not under the supervision of the Python
interpreter, \ffcv can escape the constraints of Python's global interpreter
lock (GIL) and can rely on threads instead of sub-processes like most libraries (the
GIL typically only allows a single thread to use the Python interpreter at once,
generally making multi-threading infeasible). 

Threads yields two important
advantages for \ffcv. First, threads can collaborate directly by reading/writing memory instead of
using (expensive) communication primitives. \ffcv workers can
therefore work together on same batch instead of having to work on their
own, improving on latency and saving large quantities of
memory (i.e., \ffcv's memory usage is typically constant instead of
scaling with the number of workers). 
Second, since they share the same CUDA context, all data preparation operations 
running on GPUs (e.g., data copying, augmentations) can be run asynchronously
and---critically---in parallel with respect to the training loop, reducing the
length of the critical path. See Appendix~\ref{app:multithreading_vs_multiprocessing} for more details.

\section{Case Studies}
\label{sec:eval}
\vskip -.2cm
In this section, we showcase \ffcv's versatility by illustrating how it can
dramatically accelerate model training in three common practical settings.
While one can use \ffcv with any task or modality, we center our first three use
cases around the ImageNet
ILSVRC-2012 image classification task, which comprises 1.3 million labeled training
images corresponding to 1,000 different classes. ImageNet is a standard dataset in both image
classification~\citep{russakovsky2015imagenet,he2015deep,krizhevsky2012imagenet} and
model training speed
benchmarks~\citep{mattson2020mlperf,coleman2017dawnbench,coleman2019analysis};
indeed, searching  ``ImageNet PyTorch'' on GitHub returns hundreds of thousands of repositories.

\noindent We show that \ffcv
enables dramatic speedups over typical setups\footnote{
    In this report, we compare FFCV to the standard PyTorch dataloader
    rather than specialized solutions like NVIDIA DALI (which tend to be 
    less versatile and significantly harder to use). Our goal 
    is to illustrate how a simple drop-in replacement enables greatly accelerated 
    training in a variety of settings. For more benchmarks, see \url{docs.ffcv.io}.
} in the following cases:
\begin{itemize}[leftmargin=.5cm]
    \item {\bf Single-model training}: We first
     use \ffcv to train a ResNet-50 on ImageNet to 75\% accuracy in 20 minutes
     on a single node.
    \item {\bf Multi-model training}: We then consider the setting where a
    researcher wants to train {\em several} small models in parallel (e.g., to
    obtain confidence intervals or perform hyperparameter search). We show that
    \ffcv{} can train 8 ResNet-18s at the same time (one per GPU) without
    incurring any additional overhead over single-GPU training. This
    demonstrates that \ffcv enables both (a) efficient training of high throughput
    models and (b) low-overhead concurrent training.
    \item {\bf Low-memory training}: Finally, we consider the
    (practically common) setting in which the dataset does not fit into machine
    memory (RAM). Via process-level caching and a
    quasi-random sampling scheme (exposed to users via just two lines of code),
    \ffcv{} accelerates training even when reading data from slow disks (and
    even from networked file systems) with minimal performance overhead.
\end{itemize}
We additionally demonstrate \ffcv{}'s drop-in applicability beyond computer vision tasks:
\begin{itemize}[leftmargin=.5cm]
    \item {\bf GPU-enabled sparse regression:} 
    In just a few lines of code, \ffcv{} can considerably speed up an
    iterative SAGA \citep{defazio2014saga} solver (similar to that of
    \citet{wong2021leveraging}) by simply replacing a default PyTorch
    data loader (loading from a memory-mapped file) with an \ffcv{} loader.
\end{itemize}

\subsection{Training a single model}
\label{sec:rn50}
We first study the simplest use case of \ffcv: training a single model on
ImageNet as fast as possible. By combining the data loading speed of \ffcv{}
with known ImageNet training optimizations, we are able to establish a competitive speed/accuracy tradeoff for the benchmark task of training a
ResNet-50 on ImageNet (Figure~\ref{fig:scatterplot}).

\paragraph{Fast training.} \vskip -.2cm
We begin with an overview of the training algorithm
itself---a long line of work has explored various modifications to standard
training that have been shown to improve speed and/or accuracy, of which we use
Blurpool \citep{zhang2019making}; NoWD-BN \citep{jia2018highly};
linear learning rate annealing \citep{li2020budgeted}; test-time augmentation and
resizing \citep{touvron2019fixing}; and progressive resizing (i.e., we start
training at 160px resolution and then increase to 192px 75\% of the way through
training).

\paragraph{Fast data loading.} With \ffcv we JPEG-compress 50\% of the dataset,
compromising between compute (i.e. faster image processing, as 50\% of the
images come pre-decoded) and available memory. Doing so allows our system to:

\begin{itemize}
    \item[(a)] reap the full benefits of progressive resizing. Even at smaller
    resolutions like 160px in which the GPU has much higher throughput, we can
    still fully saturate the GPU as we are neither data reading nor processing
    bottlenecked;
    \item[(b)] outsource augmentations to the CPU. Since the dataset is cached
    (largely decoded) in memory, we can use CPU cycles for augmentations that
    would otherwise go to decoding.
\end{itemize}

\begin{figure*}
    \centering
    \begin{subfigure}{.3\linewidth}
        \centering
        \includegraphics[width=\textwidth]{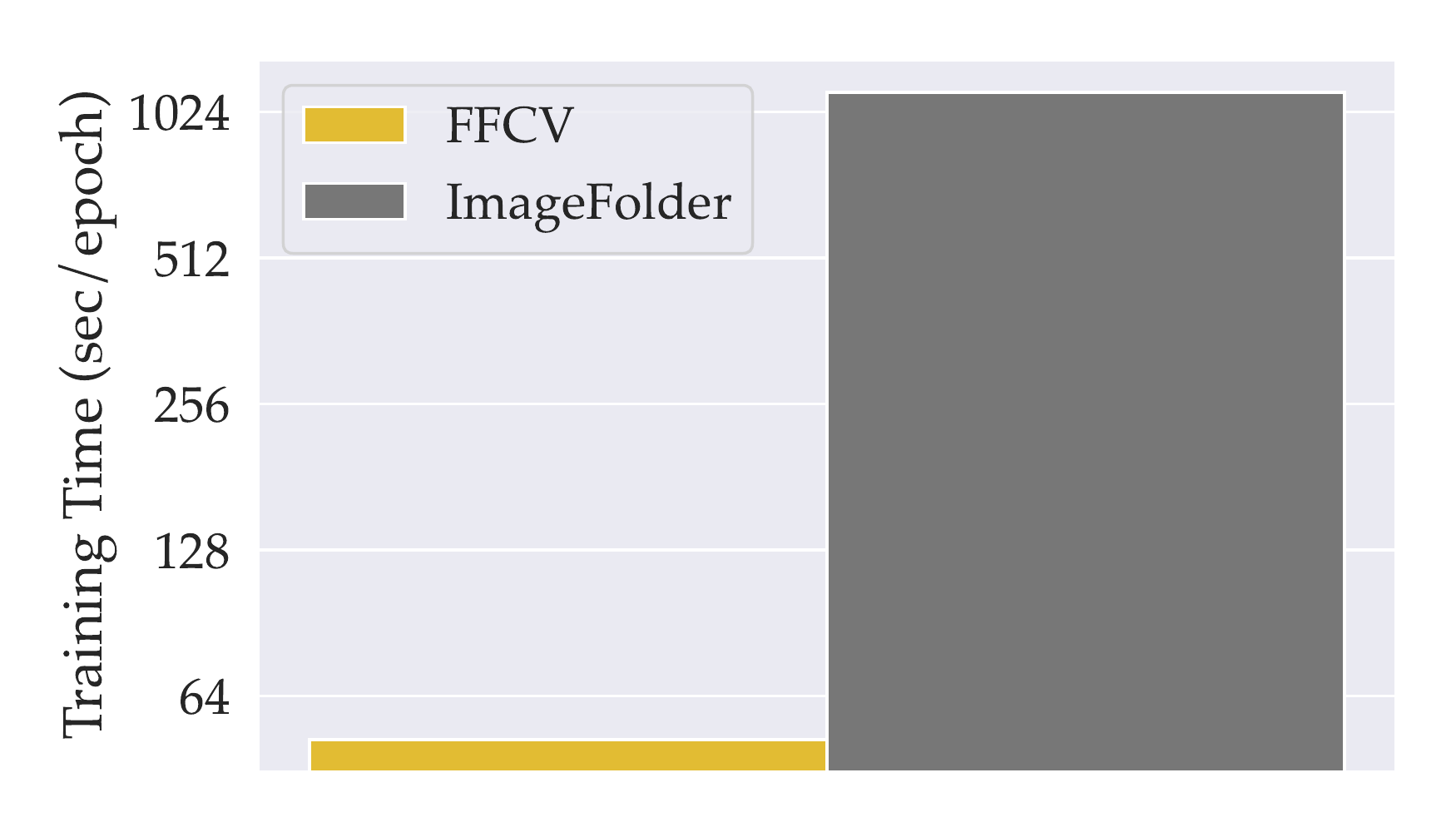}
        \caption{Single-model (1x RN-50 on 8xA100)}
        \label{fig:rn50_results}
    \end{subfigure}
    \hspace{1em}
    \begin{subfigure}{.3\linewidth}
        \centering
        \includegraphics[width=\textwidth]{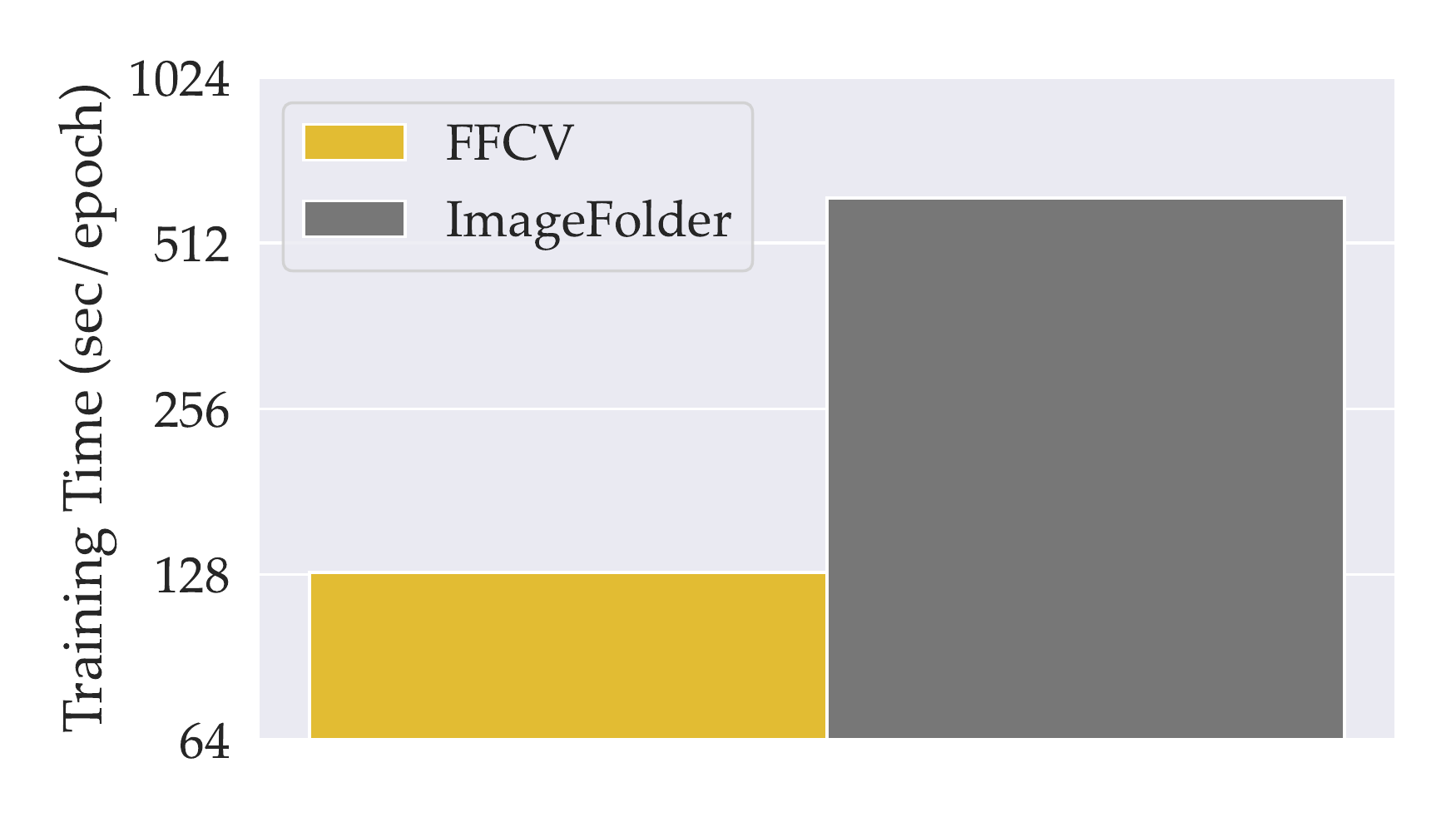}
        \caption{Multi-model (8x RN-18 on 1xA100 each)}
        \label{fig:rn18_results}
    \end{subfigure}
    \hspace{1em}
    \begin{subfigure}{.3\linewidth}
        \centering
        \includegraphics[width=\textwidth]{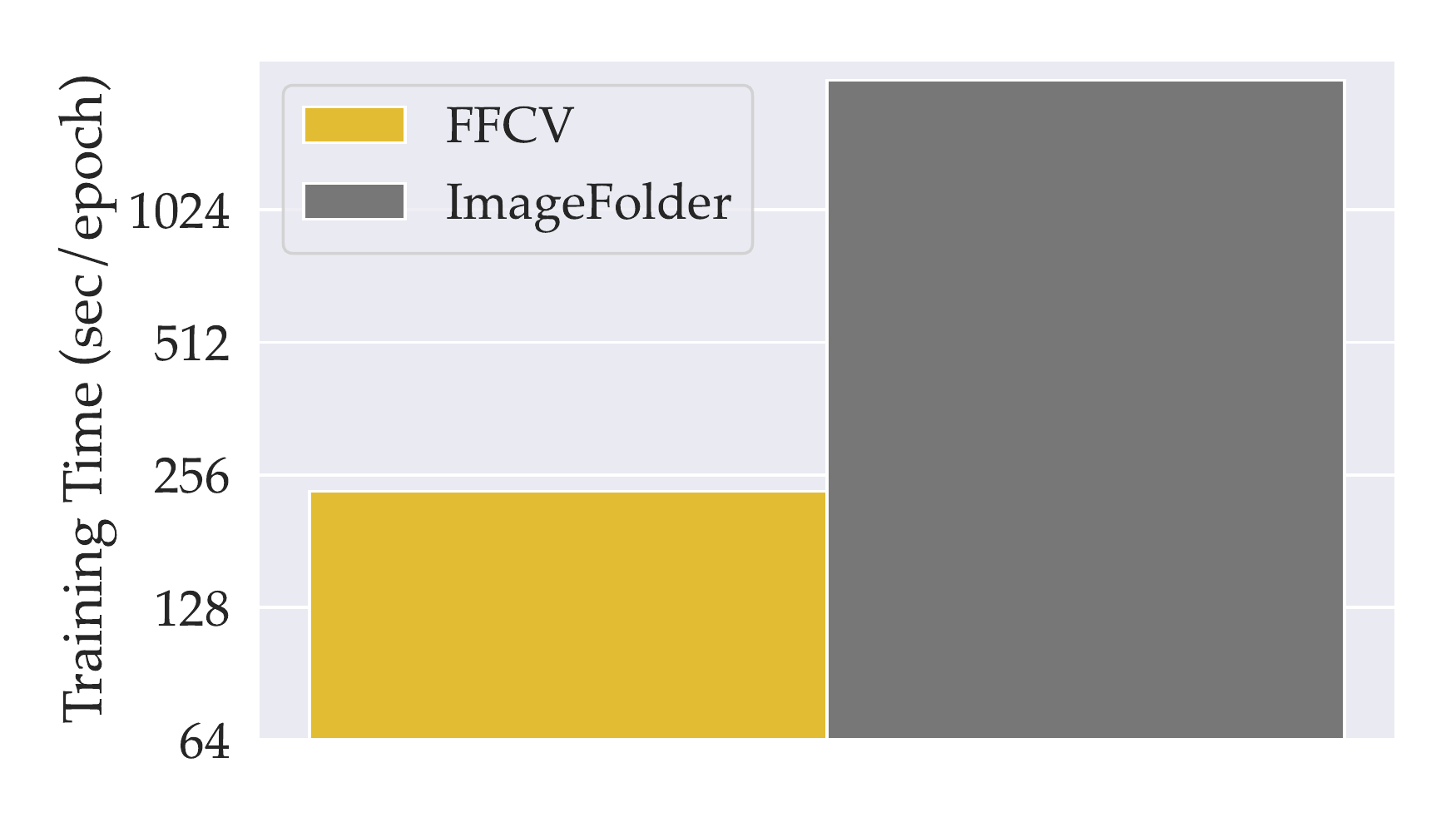}
        \caption{Low-memory (1x ResNet-18 via Network File System)}
        \label{fig:disk}
    \end{subfigure}
    \caption{A comparison between the time to train one epoch on ImageNet using \ffcv and PyTorch's ImageFolder.}
\end{figure*}

\paragraph{Evaluation.}
We compare our \ffcv-enabled optimized ImageNet example to these
baselines:
\begin{itemize}[leftmargin=.5cm]
    \item {\bf PyTorch ImageNet example:} As a naive baseline, we take the
    PyTorch ImageNet example code, which is both unoptimized and slow to load
    data---we use the reported accuracy from torchvision and multiply the
    per-epoch time (from Figure \ref{fig:rn50_results} by 90 to obtain
    an optimistic estimate of total training time. We modify the code to use half
    precision. This loader is far and away
    the most popular loader seen in open source/research implementations, and
    is used by the most popular open-source ImageNet training libraries
    \citep{wightman2019pytorch,falcon2019pytorch,developers2016pytorch}.
    \item {\bf MosaicML:} Finally, we consider models trained with the MosaicML
    Composer~\citep{team2021composer} training system (as of Februrary 2022\footnote{
        Since the release of \ffcv, the MosaicML Composer has replaced its dataloader 
        with \ffcv---for the sake of comparison, we thus use the pre-FFCV version 
        of the Composer.
    }) as a baseline. 
    MosaicML Composer supports many kinds of training options, including
    MixUp~\citep{zhang2017mixup}, specialized
    optimizers~\citep{foret2021sharpness}, squeeze-excitation
    blocks~\citep{hu2018squeeze}, and more. We compare against the
    \textit{pareto optimal} MosaicML Composer-trained models in terms of
    training speed and accuracy (\citet{team2021composer} study models trained
    across combinatorial choices of training techniques).
\end{itemize}
We run all implementations on an AWS EC2 p4d.24xlarge machine.
We report our results in Figure~\ref{fig:scatterplot};
our system obtains the best accuracy vs. speed trade-off across all
baselines. In particular, to obtain 75\% accuracy we require only
20 minutes, much faster than any of the tested baselines.

\subsection{Training multiple models}
\vskip -.2cm
Another common paradigm in machine learning is training multiple models
simultaneously. For example, we may want to perform a grid search for optimal parameters, or
rerun a model with the same training parameters to obtain confidence intervals on
results. In what follows, we show that \ffcv allows for much faster parallel
training than existing methods: \ffcv has automatic support for OS-level caching
and is high throughput enough to support even eight ResNet-18 models training 
simultaneously (ResNet-18 models
have nearly three times the throughput of ResNet-50 models since they are smaller).

\paragraph{Evaluation.} 
\vskip -.2cm
Using the same AWS EC2 p4d.24xlarge machine as above,
we run eight concurrent training routines on ResNet-18 models using \ffcv and compare with the PyTorch ImageNet example baseline (originally described in Section~\ref{sec:rn50}).

Each training routine has access to one eighth of the available vCPUs (12) and
one A100. For \ffcv, we use image datasets that have been originally
scaled to 350px, and we perform no JPEG compression, storing only image pixel
values. Our throughput results can be found in Figure~\ref{fig:rn18_results}; we find that
\ffcv{} has greater throughput than PyTorch's default ImageFolder loader,
despite not requiring any specialized hardware for decoding.

\subsection{Low-memory training}
\vskip -.2cm
In the previous two examples, we operated in a setting where the machine being
used for training has sufficient memory (RAM) to cache the entire ImageNet
dataset (in particular, our 50\% compressed version of ImageNet is 339GB).
In many scenarios, however, we do not have sufficient RAM to cache even a
fully JPEG-compressed dataset, and are thus forced to load images directly from
the filesystem. Normally, this process incurs additional significant
training cost, especially in settings where the filesystem is mounted on a
networked drive (or any other slow disk).

Here, we show that with minimal changes to existing code, \ffcv{} enables fast
training even in such resource-constrained setups. By changing
only two lines of code, we can enable {\em process-level caching} and {\em
quasi-random loading}. These optimizations
together ensure that read-constrained, memory-constrained systems operate at
as high a throughput as possible; see Section~\ref{sec:implementation} for details.

\paragraph{Evaluation.} Just as in the last section, we compare to the default PyTorch loader. The results,
shown in Figure~\ref{fig:disk}, illustrate that \ffcv indeed enables faster
training in memory-limited settings.

\subsection{Beyond computer vision}
    \begin{figure}
    \centering
    \includegraphics[width=.8\textwidth]{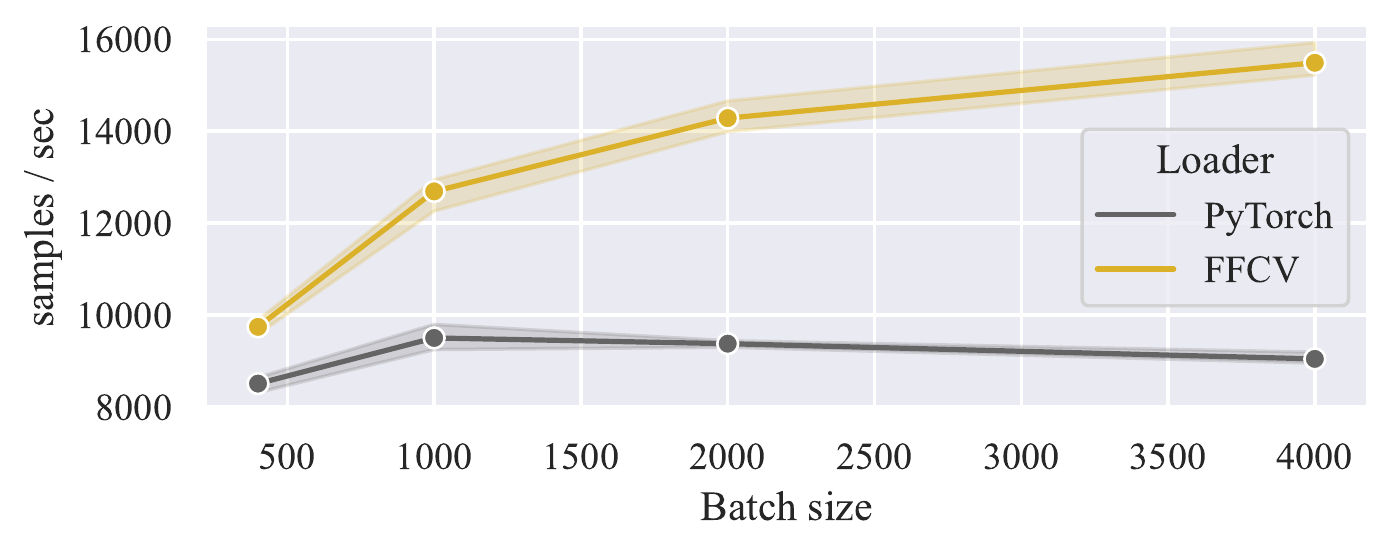}
    \caption{\ffcv{} provides significant speedups over the default PyTorch data
    loader, even for non-vision modalities. Here, we replace only the dataloader
    of a sparse regression solver with \ffcv.
    This simple change makes the solver 1.6 times faster.} 
    \label{fig:ffcv_dm_plot}
    \vskip -.2cm
    \end{figure}
Finally, we show the applicability of \ffcv beyond just computer vision
workloads. Specifically, we consider a large-scale sparse linear regression
problem with $n=100,000$ training points and dimensionality $d=50,000$.
We use an optimized iterative SAGA-based optimizer \citep{wong2021leveraging} for solving sparse linear regression problems. 
In Figure \ref{fig:ffcv_dm_plot}, we compare the unmodified code (which loads data
using the standard PyTorch data loader, reading from a memory-mapped file) with a
drop-in \ffcv replacement. The results indicate that even outside typical
computer vision setups, \ffcv is an effective drop-in replacement for
default data loaders.

\section{Related Work}
In this section, we discuss some prior work in the field of accelerating machine
learning training.

\paragraph{Data pipelines in ML.}
\citet{mohan2021analyzing} find that DNN training time is dominated by data loading in various settings.
DALI \citep{nvidia2018dali} uses custom input data pre-processing pipelines, with the option of off-loading some work to the GPU.
\citep{aizman2019high} introduce AIStore, a storage system, and WebDataset, a storage format based on POSIX tar, to enable high performance I/O for large scale deep learning.
\citet{murray2021tf} analyze millions of jobs on Google cloud and find that
they spend a significant fraction of time in the input pipeline; they find that
optimizing input pipeline performance is critical to end-to-end training time.
Their framework \texttt{tf.data} allows users to build and execute efficient
input pipelines, assisting with parallelism, caching, and static optimizations.
\citet{kakaraparthy2019case} find that ML experiments with concurrent jobs
(such as grid search) benefit from unifying data loading across jobs.
Alternatively, \texttt{kornia} \citep{riba2020kornia} implements
standard image processing functions for GPUs.

\paragraph{Speedups from other sources.}
While our work and those cited above focus on removing the data bottlenecks in current ML workloads,
large speed improvements also come from other sources, including hardware and
algorithmic improvements, which allow models to achieve similar accuracies with
less training. These include better architectures (e.g., ResNet
\citep{he2015deep}), optimizations (e.g., batch normalization
\citep{ioffe2015batch}, cyclic LR \citep{smith2017cyclical}), data
augmentation (e.g., MixUp \citep{zhang2017mixup}), among others.

\paragraph{Fast training on ImageNet.}
Our evaluation focuses on fast training on ImageNet, a standard in model training speed
benchmarks~\citep{mattson2020mlperf,coleman2017dawnbench,coleman2019analysis}. Many prior works \citep{goyal2017accurate,jia2018highly,you2017100,akiba2017extremely}
use distributed training with extremely large batch sizes 
to reduce training time.
Beyond the increase in engineering complexity arising from distributed training,
training models with large batch sizes comes with its own challenges, for instance, proper tuning of the learning rate \cite{dettmers2019sparse,you2017100}.
Moreover, extreme resource usage, including large communication overheads necessitated by distributed training
\cite{coleman2019analysis}, reduce their usability.

\section{Conclusion}
In this work, we present \ffcv, an optimized framework for
eliminating data bottlenecks in machine learning model training routines. We use
\ffcv to substantially improve speed/accuracy tradeoff for the ImageNet dataset,
and demonstrate (through a series of case studies) the potential for \ffcv to
speed up almost any ML workload. The main limitation of our work is that it does
not address non-data related bottlenecks in training, and might thus yield less
significant (but still non-zero) improvements in settings where data is fast to
load and process (e.g., natural language processing) or where models are very
large and dominate training time (e.g., large pre-trained vision models).

\section{Acknowledgements}
Work supported in part by the NSF grants CNS-1815221 and DMS-2134108, and Open
Philanthropy. This material is based upon work supported by the Defense Advanced
Research Projects Agency (DARPA) under Contract No. HR001120C0015.

\clearpage
\printbibliography

\clearpage
\clearpage
\onecolumn
\appendix

\section{Quasi-Randomness vs. Webdataset Sharding}
\label{app:quasirand_vs_shards}
In Section~\ref{sec:challenge_2}, we mentioned that, while \ffcv's quasi-random loading strategy is similar to WebDataset's sharding, the two strategies differ in key aspects. Here we discuss in more detail why this is the case. 

In systems like Webdataset, sharding is done to reduce the number of files and relieve some strain on the file system. However, each file system has a block size after which random reads become sequential reads. In almost all system, this lies below 2MB. So, if Webdataset was to create shards of 2MB, it would create a dataset with way too many files that it overwhelms the file system.

On the other hand, \ffcv's .beton format (by design) decouples this phenomenon by {\em sharding internally a single large file}. This means that our dataset is a single file the is structured to imitate sharding, as shown in Figure~\ref{fig:beton_imagenet_example}. This enables us to increase the quality of randomness as our shards are typically orders of magnitude smaller than WebDataset’s. At the same time, we avoid overwhelming the filesystem as long as we pick the minimum shard size that will satisfy the file system while retaining a single file.

\section{Why does \ffcv Use Multithreading Instead of Multiprocessing?}
\label{app:multithreading_vs_multiprocessing}
In this section, we discuss in more detail (than Section~\ref{sec:challenge_4}) why \ffcv relies to threads instead of sub-processes.

While many libraries, such as PyTorch, use multiprocessing during data loading to avoid the GIL, this comes with a large performance drop. In such strategies, all workers have to report back to the main process (running on a single thread). This often leads to lower performance with very large worker pools than with smaller as the main thread becomes overwhelmed. To avoid this, we resorted in \ffcv to multi-threading instead, thus enabling each thread to write an individual sample directly in place in RAM without any inter-process communication. This also enables multiple threads to work simultaneously on the same batch, and avoids situations like when PyTorch's main thread could be sitting idle waiting for any of its workers to have their batches ready, although if these workers could accumlate their work (which is what we do in \ffcv), the main thread would have not have been idle.

\section{Comparison of JIT Compilers Alternatives}
\label{app:compiler_comparison}
We motivate our decision to use Numba to compile FFCV's pipelines by
elaborating on the pros and cons of potential alternatives:

\paragraph{\texttt{torch.script}:}

This JIT compiler, included as part of \texttt{Pytorch} was designed to
optimize inference speed of deep neural networks and ease the transition
between research code and deployment environments. In this context, it makes
sense for it to only support the most basic python features and the Pytorch
library itself. Relying on this solution for FFCV would have mean giving up on
data augmentations written in \texttt{numpy} and interoperability with other
deep learning frameworks. While \texttt{torch.scrip} can generate very fast
code for some ML oriented operations (e.g matrix multiplications), it lacks
some optimizations (\texttt{for} loops) and does not release the \texttt{GIL}
which makes would have made it particularly unsuited for FFCV.

It is still important to note that it still possible to leverage
\texttt{torch.script} within FFCV: users are allowed to introduce in their
pipeline functions compiled with it as long as they operate on the GPU. Indeed,
FFCV since does not compile those, it does not interacting with Numba and could
even bring additional performance improvements in some scenarios.

\paragraph{\texttt{TVM}:}

TVM\footnote{\url{https://tvm.apache.org/}} shares many properties with Numba.
They both are capable to generate LLVM IR, and can leverage the same
optimization passes. Their performance should in theory be very similar. TVM
has other advantages like being able to target other environments, this would
not be useful for FFCV. The main differentiating factor remaining is the API.
The familiar numpy augmented python seemed to be the most approachable for
potential users. On the other end, TVM which was designed to represent and
optimize complex neural network architectures  would have been much more
cumbersome to work with in this scenario.

\section{Omitted Figures}
\label{app:ommited}

\begin{figure*}[!h]
    \centering
    \includegraphics[width=.8\textwidth]{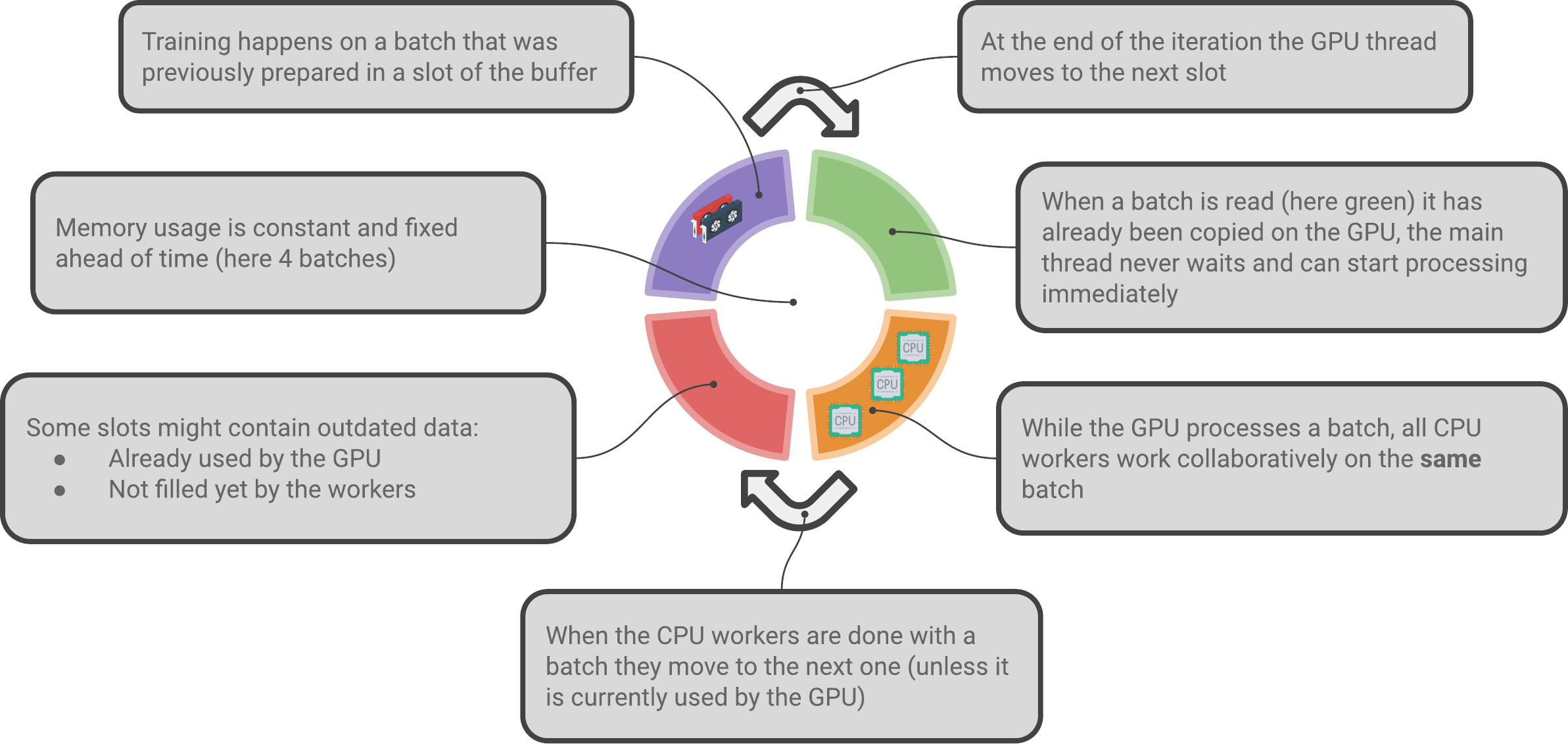}
    \caption{\label{fig:cyclical_buffer} Illustration of \ffcv's circular buffer. The
    producer (data processing pipeline) works on an entry of the buffer while the consumer
    (user's code) uses a previously filled one. When done, they moving clockwise
    and pause when they encounter each other.
    }
\end{figure*}

\end{document}